\documentclass[letterpaper, 10 pt, conference]{ieeeconf}  

\IEEEoverridecommandlockouts                              

\usepackage{graphics} 
\usepackage{epsfig}
\usepackage{times} 
\usepackage{amsmath} 
\usepackage{amssymb}  

\usepackage{xcolor}

\usepackage{bm}
\usepackage{cite}
\usepackage{tabularx}
\usepackage{multirow}
\usepackage{booktabs}
\usepackage[ruled,linesnumbered]{algorithm2e}
\usepackage{float}
\usepackage{url}

\begin{document}

\title{\LARGE \bf
CIG-RL: Curiosity-Driven Information-Guided Reinforcement Learning for Source Term Estimation in Uncertain Environments}

\author{Junhee Lee$^{1}$, Seunghwan Kim$^{2}$, Hongro Jang$^{1}$, Hyungjin Kim$^{1}$, \\ Hyoungho Park$^{2}$, Changseung Kim$^{2}$ and Hyondong Oh$^{1}$%
\thanks{$^{1}$J. Lee, H. Jang, H. Kim and H. Oh are with the Department of Mechanical Engineering, Korea Advanced Institute of Science and Technology (KAIST), Daejeon 34141, Republic of Korea.
(Corresponding author: Hyondong Oh)
{\tt\footnotesize \{ljh0124, hlomk, gudwls124z, h.oh\}@kaist.ac.kr}}
\thanks{$^{2}$S. Kim, H. Park and C. Kim are with the Department of Mechanical Engineering, Ulsan National Institute of Science and Technology (UNIST), Ulsan 44919, Republic of Korea.
{\tt\footnotesize \{kevin6960, gudgh1630, pon02124\}unist.ac.kr}}
\thanks{This work has been submitted to the IEEE for possible publication.
Copyright may be transferred without notice, after which this version
may no longer be accessible.}
}

\markboth{}
{}

\maketitle

\maketitle
\thispagestyle{empty}
\pagestyle{empty}

\begin{abstract}
Source term estimation (STE), which aims to estimate key properties of the gas source, is essential for identifying hazardous gas releases. Information-theoretic approaches have been adopted for autonomous STE using mobile sensors due to robustness in noisy environments, yet their online action selection incurs substantial computational cost. Deep reinforcement learning (DRL) provides a promising alternative with its fast decision-making capability. In DRL-based STE, the agent selects actions based on belief states of the source term updated from noisy measurement sequences. However, existing methods rely on random exploration or solely on belief uncertainty reduction without an effective exploration strategy in DRL, which can limit policy robustness in noisy environments. To address this, we propose a curiosity-driven information-guided reinforcement learning for robust and efficient STE. The proposed method promotes active exploration of novel belief state transitions that have not been sufficiently explored during training. We further introduce an uncertainty-adaptive active perception reward to guide efficient source search under uncertainty. Simulations under high-noise conditions and real-world experiments demonstrate the robustness and feasibility of the proposed framework, highlighting its potential for practical STE problems.
\end{abstract}

\section{INTRODUCTION}
\label{sec:introduction}
Hazardous gas leakage poses serious risks to human health, requiring rapid identification of the gas source properties. Estimating key properties (e.g., source location and release strength) is referred to as a source term estimation (STE) problem. STE problems are challenging because gas leakage is often invisible and strongly affected by turbulent atmospheric conditions. Since it is dangerous for humans to directly search for the gas source, autonomous STE strategies using mobile sensors have attracted considerable attention~\cite{Oh}.

One of the representative strategies for STE is the information-theoretic approach, which employs Bayesian inference to estimate the source term and utilizes information theory to select actions that maximize the information gain~\cite{vergassola2007infotaxis, ristic2016cognitive, hutchinson2018entrotaxis, dualmode}.  Although this approach is effective in addressing noisy measurements, it requires computing the expected entropy reduction for all action candidates, resulting in high computational cost. Consequently, this computational cost limits planning over long horizons.

Deep reinforcement learning (DRL) offers a promising solution for addressing these issues in STE. DRL can learn search policies offline and execute them efficiently during deployment, enabling fast decision-making and non-myopic search behavior~\cite{RL_robotics, yang2025}. Consequently, many studies have adopted end-to-end DRL approaches that focus on rapid source search~\cite{hu2019plumetracing, li2024GRUPPO, singh2023emergent, ddqn_indoor}. However, these methods do not use explicit source estimators and thus cannot easily determine when the source search is complete or reliably estimate the source location.
\begin{figure}[!t]
\vspace{6pt}
\centering
\includegraphics[width=\columnwidth]{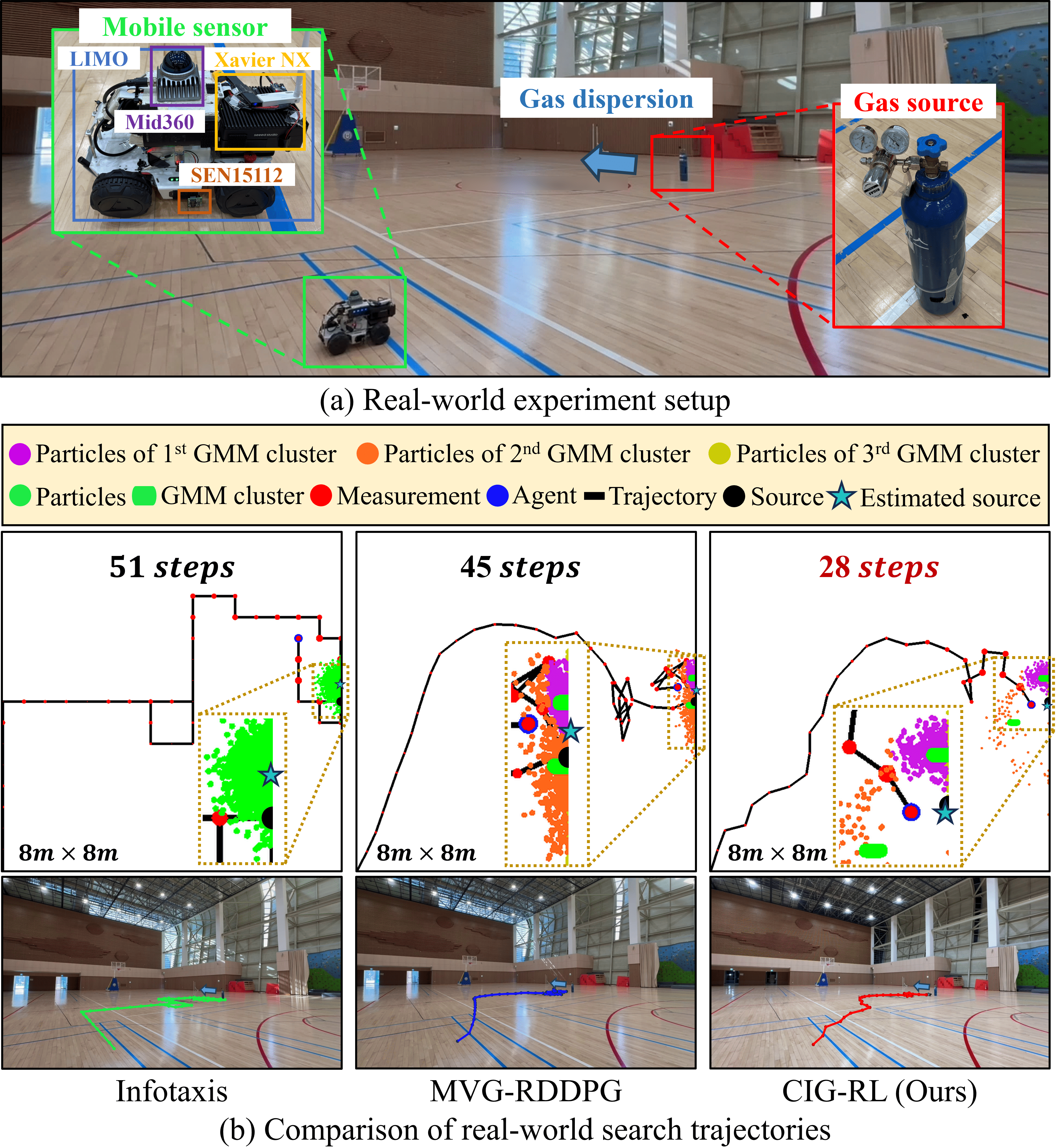}
\vspace{-20pt}
\caption{Overview of the real-world validation: (a) experimental setup with the mobile sensor and gas source; and (b) comparison of real-world search trajectories of Infotaxis, MVG-RDDPG, and CIG-RL.}
\label{experiment1}
\end{figure}

To address this limitation, recent studies have integrated source estimators into DRL-based STE frameworks~\cite{zhao2022pcdqn, park2022mvgrddpg, aaai_ste, li2023aidrl, lee2025enhanced}. In these frameworks, the source estimator probabilistically estimates the source location, typically using particle filters. Nevertheless, developing a robust search policy under highly noisy conditions remains a significant challenge. In DRL-based STE, the agent makes decisions based on particle filter-based belief states updated from noisy measurement sequences. Under severe noise, these belief states can become biased toward incorrect source hypotheses. Therefore, effective exploration of how the belief state changes with respect to time (i.e., belief state transitions) is crucial during training. However, existing methods rely on random exploration~\cite{zhao2022pcdqn, park2022mvgrddpg, aaai_ste}, which lacks a mechanism for effectively acquiring informative observations in noisy environments. To mitigate this, recent studies have introduced information gain into policy learning~\cite{li2023aidrl, lee2025enhanced}. These methods primarily guide the policy by reducing uncertainty in the current belief. However, because the belief can be distorted by noisy measurements, uncertainty reduction alone without an effective exploration may be insufficient for robust policy learning. Beyond these limitations, real-world validation of DRL-based STE methods with explicit source estimation has rarely been reported, leaving their practical robustness insufficiently verified.

In this paper, we propose a DRL-based framework that enables robust source search in highly noisy environments. The key idea is to improve policy robustness by encouraging exploration based on the novelty of belief transitions, rather than relying on random exploration~\cite{zhao2022pcdqn, park2022mvgrddpg, aaai_ste} or solely reducing current belief uncertainty~\cite{li2023aidrl, lee2025enhanced}. To this end, we develop an exploration module by leveraging a curiosity network~\cite{pathak2017curiosity}. Our exploration module predicts the next latent belief from the current latent belief, agent position, and action, and uses the discrepancy between the predicted and actual next latent beliefs as an exploration signal (i.e., auxiliary reward). This encourages the agent to explore belief transitions that remain difficult to predict from prior training experience, promoting more robust policy learning. To prevent excessive exploration while reliably guiding the agent toward the source, we further propose an uncertainty-adaptive active perception reward. This reward adaptively balances information gathering (i.e., exploration) and source-directed approach (i.e., exploitation) according to the level of the particle filter variance, enabling efficient search under noisy conditions. To validate the effectiveness and feasibility of the proposed framework, we conduct simulations in highly noisy environments and real-world experiments with $CO_2$ leakages. The main contributions are:

\begin{enumerate}
    \item A curiosity-driven belief space exploration framework for DRL-based STE, which encourages the agent to actively explore novel belief state transitions for robust source search;
    \item An uncertainty-adaptive active perception reward that integrates mutual information with particle filter variance, enhancing robustness to noisy conditions while ensuring efficient source search; and
    \item Validation in challenging high-noise simulations and real-world experiments with $CO_2$ leakage scenarios, demonstrating feasibility and practical applicability.
		
\end{enumerate}

\section{Related Work}
\label{sec:related works}
\subsection{Information-Theoretic Approaches}
Information-theoretic approaches have gained attention for their effectiveness in handling noisy gas measurements. Infotaxis~\cite{vergassola2007infotaxis}, cognitive search~\cite{ristic2016cognitive}, and dual-mode planner~\cite{dualmode} determine actions that maximize the reduction in the entropy of the source term, whereas Entrotaxis~\cite{hutchinson2018entrotaxis} selects actions that maximize the entropy of the future measurement distribution over all action candidates at each step. However, the computational cost of uncertainty calculation for action candidates can limit real-time implementation. 
\subsection{End-to-End DRL}
Recent studies have adopted DRL due to its real-time applicability and ability to learn efficient source search policies. In end-to-end DRL approaches, most studies employ memory networks to improve source search performance. Hu et al.~\cite{hu2019plumetracing} propose a long short-term memory (LSTM)-based plume-tracing framework, while Li et al.~\cite{li2024GRUPPO} introduce proximal policy optimization based on a gated recurrent unit (GRU-PPO) approach to effectively leverage historical measurements. Singh et al.~\cite{singh2023emergent} further develop a recurrent neural network agent that robustly tracks turbulent gas plumes under varying wind conditions. Beyond memory-based approaches, He et al.~\cite{ddqn_indoor} introduce a dueling deep Q-network (Dueling DQN) method using computational fluid dynamics (CFD) simulations to reflect more realistic environments. However, because these approaches do not include an explicit source estimator, they cannot reliably determine when the search is complete or accurately identify the source location.

\subsection{DRL with Explicit Source Estimators}
Unlike end-to-end DRL approaches, several studies incorporate explicit source estimators into DRL frameworks to enable more reliable source search. PC-DQN~\cite{zhao2022pcdqn} employs density-based spatial clustering of applications with noise (DBSCAN) to extract belief features from the particle filter, while MVG-RDDPG~\cite{park2022mvgrddpg} improves belief representation using Gaussian mixture model (GMM)-based feature extraction and leverages historical measurements through a GRU. AGDC~\cite{aaai_ste} further introduces an autonomous goal detection mechanism that adopts the convergence of the particle filter to effectively determine search termination. Although these methods improve belief representation or search termination, their random exploration is sample-inefficient and lacks guidance on how to explore informative states in the belief space under uncertainty. Furthermore, their methods are trained under low environmental noise conditions, which can limit robustness in real-world scenarios.

AID-RL~\cite{li2023aidrl} and GMM IG-RL~\cite{lee2025enhanced} attempt to handle this issue by utilizing information-theoretic principles in DRL. AID-RL~\cite{li2023aidrl} combines greedy exploitation toward high-concentration regions with information-directed exploration. However, its concentration-based exploitation can make the policy vulnerable to spurious high-concentration measurements in highly noisy environments. More recently, GMM IG-RL~\cite{lee2025enhanced} improves search robustness by designing GMM information gain rewards that reduce current belief uncertainty. Nevertheless, since belief can be distorted by noisy measurements, relying primarily on uncertainty reduction without an effective exploration may limit policy robustness.

\section{Problem Statement}
It is assumed that a hazardous gas source located at $\mathbf{p}_s=\left[x_s, y_s\right]^T$ emits gas particles with a release strength \( q_s \). The source term vector is parameterized as $\theta_s = \left[\mathbf{p}_s^T, {q_s} \right]^T$, and this study focuses on localizing $\mathbf{p}_s$. At each time step, the agent equipped with a gas sensor acquires gas measurements at its position \( \mathbf{p}_t = [x_t, y_t]^T \) and updates its estimate of the source term distribution using a particle filter, based on the predefined gas dispersion and sensor models. Leveraging the estimated source term, CIG-RL is applied to find the best source search policy under uncertainty. 

\subsection{Gas Dispersion Model}
For the gas dispersion model, we use the isotropic plume model~\cite{vergassola2007infotaxis}. In this model, gas particles diffuse with a particle life time \( \tau \) and diffusivity \( D \), and are advected by an average wind speed \( V \) with direction \( \chi \). The mean gas concentration $C(\mathbf{p}_t | \theta_s)$ acquired at the sensing position \( \mathbf{p}_t \) is defined as:
\begin{align}
C(\mathbf{p}_t|\theta_s) &= \frac{q_s}{4\pi D\,|\mathbf{p}_t - \mathbf{p}_s|} 
\exp\frac{{V}(x_t - x_s)\sin\chi}{2D} \nonumber \\
&\quad \cdot \exp\frac{-{V}(y_t - y_s)\cos\chi}{2D} 
\cdot \exp\frac{-|\mathbf{p}_t - \mathbf{p}_s|}{\lambda},
\label{eqn:concentration_model}
\end{align}
\noindent where $\lambda = \sqrt{D\tau / \left(1 + \frac{V^2\tau}{4D}\right)}$.

\subsection{Sensor Model}
In real-world scenarios, sensor measurements are affected by wind turbulence and sensor noise. To capture these effects, we adopt a Gaussian noise model as the sensor model~\cite{dualmode}. The sensor measurement \( z_t \) obtained by the agent at the sensing position \( \mathbf{p}_t \) is expressed as:
\begin{equation}
z_t = C(\mathbf{p}_t | \theta_s) + \nu_{{env}} + \nu_{{sensor}},
\label{eqn:sensor_model}
\end{equation}
\noindent where \(\nu_{{env}}\) and \(\nu_{{sensor}}\) denote the noise arising from the wind turbulence and the sensor measuring process, respectively. Both terms are assumed to follow Gaussian distributions, i.e., 
$\nu_{env} \sim \mathcal{N}(0,\sigma_{env}^2)$ and 
$\nu_{sensor} \sim \mathcal{N}(0,\sigma_{sensor}^2)$, 
where $\sigma_{env}$ represents the instability of the wind conditions. 
Here, $\sigma_{sensor} = \beta \cdot C(\mathbf{p}_t | \theta_s)$, where $\beta$ denotes the sensor noise level. 
The likelihood of the sensor measurement $z_t$ for a given source term $\theta_s$ is defined as:

\begin{equation}
p(z_t | \theta_s) = \frac{1}{\sigma_T \sqrt{2\pi}} \exp -\frac{(z_t - C(\mathbf{p}_t | \theta_s))}{2\sigma_T^2},
\label{eqn:likelihood}
\end{equation}
\noindent where $\sigma_T = \sqrt{\sigma_{\textit{}{env}}^2 + \sigma_{\textit{}{sensor}}^2}$ denotes the overall standard deviation of the total noise. 

\setlength{\abovedisplayskip}{6pt}
\setlength{\belowdisplayskip}{6pt}
\setlength{\abovedisplayshortskip}{4pt}
\setlength{\belowdisplayshortskip}{4pt}
\subsection{Particle Filter}
We utilize a particle filter to estimate the source term, as it is well-suited for handling the nonlinearity of the source term and remains robust under high levels of sensor measurement noise~\cite{particlefilter}. The source term probability distribution can be expressed by \( N_p \) particles as: 
\begin{equation}
p(\theta_{t} | z_{1:t}) = \sum_{i=1}^{N_p} w_t^i \delta(\theta_{t} - \theta_{t}^i),
\label{eqn:belief_state}
\end{equation}
\noindent where \( w_t^i \) indicates the normalized weight of each particle, \( \delta(\cdot) \) is the Dirac delta function, and \( \theta_{t}^i \) denotes each sampled particle. Upon receiving a new sensor measurement, the unnormalized particle weight can be sequentially updated as:

\begin{equation}
\widetilde{w}_{t+1}^i = p(z_{t+1} | \theta_{t}^i) \cdot w_t^i.
\label{eqn:normalized_weight_update}
\end{equation}
The measurement likelihood \( p(z_{t+1} |\theta_{t}^i) \) is computed using the defined gas dispersion and sensor models given in~\eqref{eqn:concentration_model} and~\eqref{eqn:likelihood}, respectively.
Then, the normalized particle weight \( w_{t+1}^i \) is calculated as:
\begin{equation}
w_{t+1}^i = \frac{\widetilde{w}_{t+1}^i }{\sum_{i=1}^{N_p} \widetilde{w}_{t+1}^i }.
\label{eqn:normalized_weight}
\end{equation}

Since the particle filter can suffer from the degeneracy, in which most particle weights collapse toward zero, resampling is applied. After resampling, the Markov chain Monte Carlo (MCMC) method~\cite{resampling} is employed to improve particle impoverishment.

\begin{figure*}[t!]
    \centering
    \includegraphics[width=0.7607\linewidth]{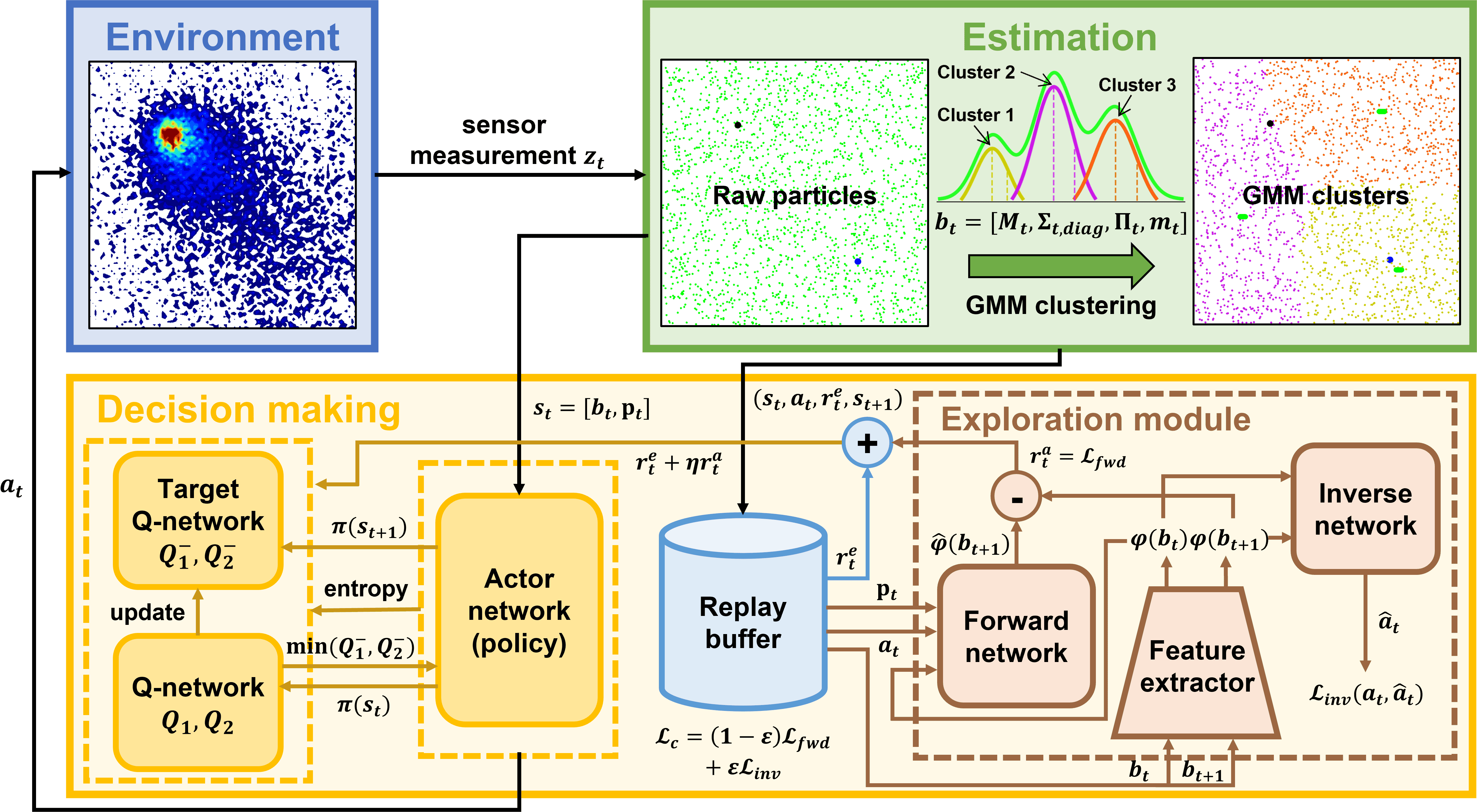} 
    \vspace{-7pt}
    \caption{System architecture of CIG-RL. The exploration module predicts the next latent belief representation and uses the discrepancy between the predicted and actual latent belief representations as an auxiliary reward, encouraging exploration of insufficiently learned belief transitions. This reward is combined with the uncertainty-adaptive active perception reward and used to update the critic network, enabling robust and efficient policy learning.}
    \label{CIG_SAC_rev}
\end{figure*}

\section{Curiosity-Driven Information-Guided Reinforcement Learning}
In this section, we introduce a curiosity-driven information-guided reinforcement learning framework to achieve robust source search in noisy environments. State and action are first introduced, and the curiosity-driven belief space exploration for STE is explained. Then, the uncertainty-adaptive active perception reward is presented, and the overall learning framework is outlined. 

\subsection{State and Action} 
\subsubsection{State}
To enhance the efficiency of source estimation, Gaussian mixture model (GMM) clustering~\cite{GMM} is applied to extract features from the particle filter. The belief state is approximated using a GMM, characterized by the mean of Gaussian distribution $\bm{M}_t = [\mu_t^{1}, \mu_t^{2}, \ldots, \mu_t^{L}]$, its covariance $\boldsymbol{\Sigma}_t = [\Sigma_t^{1}, \Sigma_t^{2}, \ldots, \Sigma_t^{L}]$, and the mixing weights $\boldsymbol{\Pi}_t = [\pi_t^{1}, \pi_t^{2}, \ldots, \pi_t^{L}]$, where $L$ denotes the number of GMM clusters. Here, $L$ is set to 3. To avoid GMM label-switching ambiguity, the GMM components are consistently ordered. The GMM probability of each particle is calculated as: 
\begin{equation}
p(\theta_t^i | \bm{M}_t, \boldsymbol{\Sigma}_t, \boldsymbol{\Pi}_t) = \sum_{l=1}^{L} \pi_t^l \mathcal{N}(\theta_t^i | \mu_t^{l}, \Sigma_t^l),
\label{eq:GMM_probability}
\end{equation}
\noindent where $\mathcal{N}$ denotes the probability density function of the $l$-th GMM component. The state is then defined as:
\begin{equation}
{s}_t = \left[b_t, \mathbf{p}_t \right],
\label{eq:state_definition}
\end{equation}
\noindent where $b_t = [\bm{M}_t, \boldsymbol{\Sigma}_{t,\mathit{diag}}, \boldsymbol{\Pi}_t, m_t]$ denotes the belief state, $\boldsymbol{\Sigma}_{t,\mathit{diag}} = [\mathit{diag}(\Sigma_t^{1}), \mathit{diag}(\Sigma_t^{2}), \ldots, \mathit{diag}(\Sigma_t^{L})]$ contains the diagonal terms of the GMM component covariance matrices, and $m_t$ represents the mean of all particles.

\subsubsection{Action}
The action space is parameterized by the heading direction \(a_t\), while the movement distance is fixed at \(\Delta d\) for each step. The next sensing position is given by:
\begin{equation}
\mathbf{p}_{t+1} = \mathbf{p}_t + 
\begin{bmatrix}
\cos(a_t) \\
\sin(a_t)
\end{bmatrix} \cdot \Delta d,
\end{equation}

\subsection{Curiosity-Driven Belief Space Exploration for STE}
As discussed earlier, policy learning in STE depends on observing informative measurement sequences under uncertainty, making an effective exploration strategy essential. Particularly, since noisy measurements can mislead the belief, the agent must sufficiently and efficiently experience belief state transitions during training. To this end, we enhance exploration by introducing a curiosity-based learning signal, inspired by~\cite{pathak2017curiosity}. Our approach formulates curiosity in the belief space using particle filter-based belief features, where a high prediction error serves as an indicator of potentially novel or insufficiently learned belief transitions. 
\begin{figure}[t]
\centering
\includegraphics[width=0.4352\textwidth]{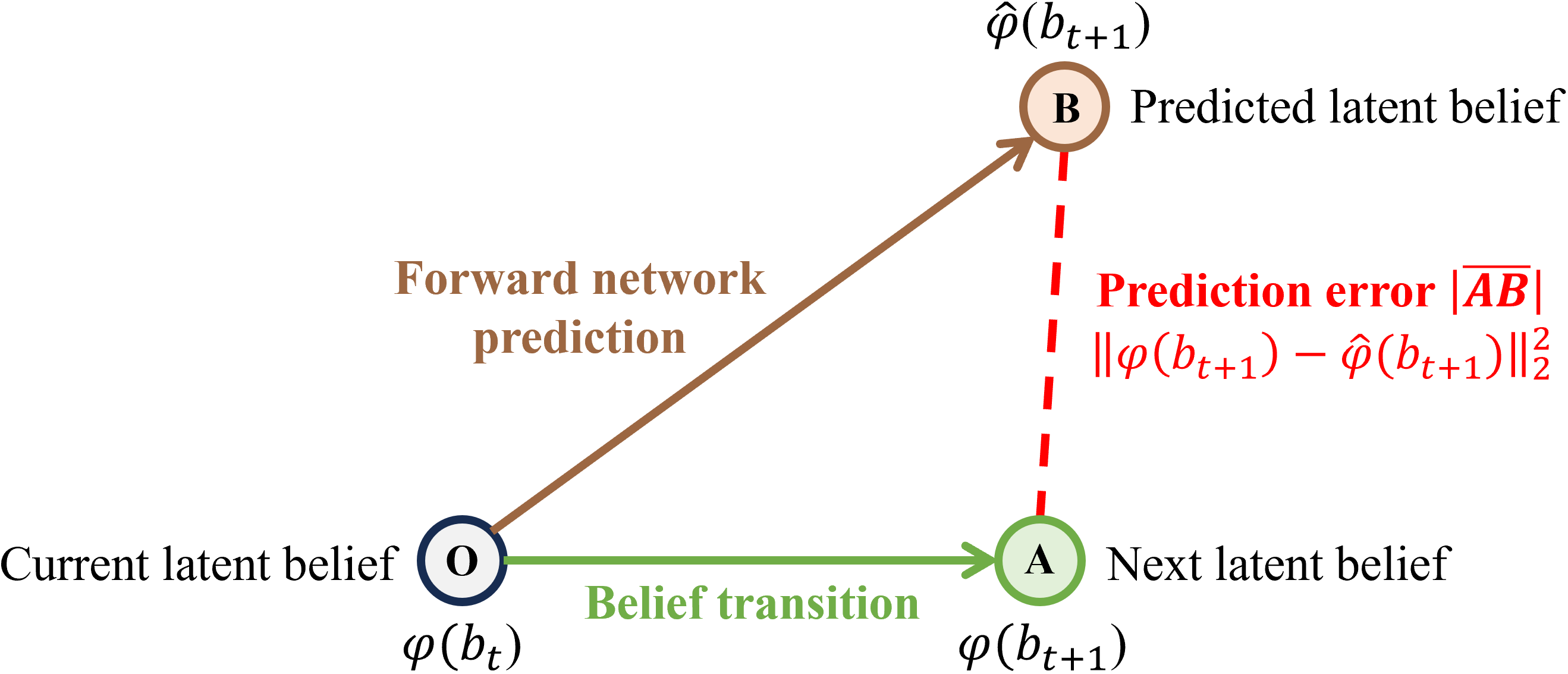}
\vspace{-8pt}
\caption{Prediction error between the next latent belief representation and the predicted latent belief representation in curiosity-driven belief space exploration for STE problems.}
\label{STE_curiosity_rev}
\end{figure}

As shown in Fig.~\ref{CIG_SAC_rev}, we design a curiosity network as an exploration module composed of a feature extractor, a forward network, and an inverse network, parameterized by \( \xi_{\mathit{feat}}, \xi_{\mathit{fwd}}, \) and \(\xi_{\mathit{inv}} \), respectively. The feature extractor encodes the particle filter-based belief state into a latent belief representation. Given the consecutive belief states $b_t$ and $b_{t+1}$, the latent belief representations are obtained as \(\varphi(b_t)\) and \(\varphi(b_{t+1})\), respectively. The forward network predicts the next latent belief representation $\hat{\varphi}(b_{t+1})$ from $\varphi(b_t)$, $\mathbf{p}_t$, and $a_t$. The prediction error between \(\varphi(b_{t+1})\) and $\hat{\varphi}(b_{t+1})$ is used as the forward network loss:
\begin{align}
\mathcal{L}_{\mathit{fwd}} 
&= \left\lVert \varphi(b_{t+1}) - \hat{\varphi}(b_{t+1}) \right\rVert_2^2.
\end{align}
To ensure that the latent representation retains meaningful action-relevant belief in the forward network prediction, we also employ an inverse network. The inverse network predicts the action \(\hat{a}_t\) that caused the transition between consecutive belief states given \(\varphi(b_t)\) and \(\varphi(b_{t+1})\), and is trained with the following loss function:
\begin{align}
\mathcal{L}_{\mathit{inv}} &= \mathrm{MSE}(a_t, \hat{a}_t).
\end{align}
The curiosity network is trained by minimizing a weighted sum of the forward and inverse network losses, given by:
\begin{align}
\min_{\xi_{\mathit{feat}},\xi_{\mathit{fwd}},\xi_{\mathit{inv}}} \mathcal{L}_{\mathit{c}}
= (1-\varepsilon)\,\mathcal{L}_{\mathit{fwd}}
+ \varepsilon\,\mathcal{L}_{\mathit{inv}},
\label{eq:curiosity_loss}
\end{align}
\noindent where \( \varepsilon \) balances the forward and inverse loss terms.

For STE problems, note that the belief state is extracted by GMM clustering, as in~\eqref{eq:state_definition}. As illustrated in Fig.~\ref{STE_curiosity_rev}, point \(A\) represents the latent belief representation of the next belief state, while point \(B\) denotes the latent belief representation predicted by the forward network. In the early stage of training, a noticeable gap exists between point \(A\) and \(B\), representing the prediction error measured as \( |\overline{AB}| = \left\| \varphi(b_{t+1}) - \hat{\varphi}(b_{t+1}) \right\|_2^2 \). A larger \( \left|\overline{AB}\right| \) indicates an insufficiently learned belief transition and can be used as an exploration signal to encourage exploration of such transitions. Thus, a large prediction error can encourage the policy to explore belief transitions that remain difficult for the forward network to predict. Importantly, since \( \left|\overline{AB}\right| \) corresponds to the loss of the forward network, it tends to decrease as the forward network is trained. Consequently, the agent can be guided to learn in a way that gradually reduces this prediction error. Based on this insight, we adopt the forward loss, representing the prediction error, as the auxiliary reward for DRL to encourage the agent to efficiently explore novel belief transitions, as:
\begin{align}
r_t^a = \mathcal{L}_{\mathit{fwd}}.
\end{align}
\subsection{Uncertainty-Adaptive Active Perception Reward}
Although encouraging exploration in DRL to seek novel belief transitions is critical, relying heavily on exploration is insufficient to achieve an optimal policy and may delay the search process. Therefore, an additional mechanism is required to reliably and efficiently guide the agent to the source. To this end, we propose an uncertainty-adaptive active perception reward. This reward augments mutual information with a distance-based source approach term, and adaptively balances the two through the particle filter variance to enable robust search under uncertainty while facilitating rapid approach to the source.

Utilizing mutual information based on Shannon’s entropy reduction can provide an effective way to enhance robustness under noisy conditions~\cite{lee2025enhanced}. The mutual information is employed as a reward to drive the policy toward reducing the uncertainty of the estimated source term, defined as:
\begin{align}
I\!\left(\mathbf{p}_{t+1}\right) &= - \sum_{i=1}^{N_p} w_t^i \log w_t^i \nonumber \\
&\quad + \sum_{\hat{z}_{t+1}=0}^{z_{\mathit{max}}} p(\hat{z}_{t+1}| \theta_{t}) 
\left( \sum_{i=1}^{N_p} \hat{w}_{t+1}^i \log \hat{w}_{t+1}^i \right),
\label{IG}
\end{align}
\noindent where \( \hat{z}_{t+1} \) denotes all possible future measurements, $z_{\mathit{max}}$ is the maximum value of the discretized future measurements, and \(p(\hat{z}_{t+1}| \theta_{t}) \) represents the likelihood of the future measurements, computed using~\eqref{eqn:likelihood}. Here, the potential particle weight \( \hat{w}_{t+1}^i \) is updated based on~\eqref{eqn:normalized_weight_update} and~\eqref{eqn:normalized_weight}. Furthermore, to accelerate the source search, we introduce a distance term between the agent's position and the estimated source location, defined as:
\begin{equation}
d_{t} = \left| \mathbf{p}_{t} - \hat{\mathbf{p}}_{t,s} \right|,
\label{Distance}
\end{equation}
\noindent where $\hat{\mathbf{p}}_{t,s}$ denotes the current estimated source location.

Unlike~\cite{lee2025enhanced}, to enable an efficient balance between the mutual information term and the distance term, an automatic adjustment method based on the particle filter variance is proposed. The particle filter variance can serve as the indicator of the reliability in the source estimation. The variance of all particles at time step \( t \) is calculated as:
\begin{equation}
\text{Cov}(\theta_{t}) = \sum_{i=1}^{N_p} w_t^i \left( \theta_{t}^i - m_t \right)\left( \theta_{t}^i - m_t \right)^T.
\label{eqn:covariance}
\end{equation}
\noindent Then, the active perception reward function is defined as:
\begin{equation}
r_{t,\mathit{step}} = I(\mathbf{p}_{t+1}) - 
\left(\sqrt{\mathit {tr}\!\left(\operatorname{Cov}(\theta_t)\right)}\right)^{-1} d_t.
\label{eqn:reward_final}
\end{equation}
Here, to address the unit inconsistency between the source location and release strength in \(\theta_t \), each dimension is normalized prior to variance computation. For numerical stability, $\mathit{tr}\!\left(\mathrm{Cov}({\theta}_t)\right)$ is clipped with a lower bound of $(0.1)^2$.

In this method, as a high particle filter variance indicates that the belief over the estimated source term is uncertain, the mutual information term has a greater relative influence on the reward to promote uncertainty reduction in such cases. As more measurements are collected and the variance decreases, the reliability in source estimation increases. Consequently, the distance term can contribute more strongly to the reward, guiding the agent toward the estimated source location.

Finally, the total extrinsic reward function is defined as:
\begin{equation}
r_t^e =
\left\{
\begin{array}{ll}
+10, & \text{successfully find the source,} \\[6pt]
r_{t,\mathit{step}},
& \text{for each step,} \\[6pt]
-5, & \text{outside search boundary.}
\end{array}
\right.
\label{total_reward}
\end{equation}

\subsection{Learning Framework for CIG-RL}
To train a robust source search policy under uncertainty, we propose curiosity-driven information-guided reinforcement learning (CIG-RL), a learning framework using the soft actor-critic (SAC)~\cite{SAC} while formulating auxiliary learning to enhance exploration in STE. 

We design the replay buffer \( \mathit{D} \) that stores tuples of agent-environment interaction in the form \( (s_t, a_t, r_t^e, s_{t+1}) \). This design is to facilitate diverse experiences and efficient sample reuse by randomly sampling mini-batches. As depicted in Fig.~\ref{CIG_SAC_rev}, the extrinsic active perception reward $r_t^e$, which reflects the uncertainty reduction of the estimated source term, is stored and sampled from the replay buffer. In contrast, the auxiliary reward $r_t^a$, capturing novelty in the belief state transition is updated by the curiosity network rather than replayed from past transitions. This decoupled sampling strategy prevents auxiliary reward values that were high in previously explored uncertain regions from being repeatedly reused, thereby mitigating bias toward excessive exploration during later stages of training. The extrinsic active perception and auxiliary rewards are then combined to form the total reward, which is used to update the critic network consisting of the Q-network and the target Q-network, as:
\begin{align}
    r_t^{\textit{total}} = r_t^e + \eta r_t^a,
    \label{eq:total_reward_buffer}
\end{align}
\noindent where \(\eta\) is the weight parameter between the extrinsic and auxiliary rewards. Based on the combined extrinsic and auxiliary rewards, the target Q-value is computed as:

\vspace{-4mm}  
\begin{equation}
\begin{aligned}
y_t &= r_t^{\textit{total}} \\[-2pt]
    &\; + \gamma \, \mathbb{E}_{a_{t+1} \sim \pi_\psi} 
      \left[ Q_{\bar{\phi}}(s_{t+1}, a_{t+1}) 
      - \alpha \log \pi_\psi(a_{t+1} | s_{t+1}) \right],
\end{aligned}
\label{eq:c_sac_target_q}
\end{equation}
\noindent where \( \bar{\phi} \) denotes the parameter of the target Q-network, \( \psi \) represents the parameter of the actor network, and \( \alpha \) is the temperature parameter which balances the entropy term against returns. The critic network is then trained by minimizing the following loss function:
{\setlength{\abovedisplayskip}{4pt}
\setlength{\belowdisplayskip}{4pt}
\begin{align}
\mathcal{L}_{\textit{critic}}(\phi) = \mathbb{E}_{(s_t, a_t) \sim \mathit{D}} \left[ \left(Q_\phi(s_t, a_t) - y_t \right)^2 \right].
\label{eq:critic_loss}
\end{align}
}
\noindent To optimize the policy, the actor network is trained by minimizing the following:
{\setlength{\abovedisplayskip}{4pt}
\setlength{\belowdisplayskip}{4pt}
\begin{align}
\mathcal{L}_{\textit{actor}}(\psi) = 
\mathbb{E}_{s_t \sim \mathit{D},\, a_t \sim \pi_\psi} \left[ 
\alpha \log \pi_\psi(a_t|s_t) - Q_\phi(s_t, a_t) 
\right].
\label{eq:actor_loss}
\end{align}
}
\noindent Finally, we adopt automatic entropy tuning to effectively adjust the temperature parameter \( \alpha \). The objective function for tuning \( \alpha \) can be written as:
\begin{align}
\mathcal{L}_\alpha = 
\mathbb{E}_{a_t \sim \pi_\psi} \left[ 
- \alpha \log \pi_\psi(a_t|s_t) + \widetilde{\mathcal{H}}
\right],
\label{eq:alpha_loss}
\end{align}
\noindent where \( \widetilde{\mathcal{H}} \) is the target entropy which determines the desired level of policy stochasticity. 

\begin{table}[!t]
\setlength{\abovecaptionskip}{0.1pt}
\caption{Parameter values of environments 1 and 2}
\label{tab:parameters2}
\centering
\setlength{\tabcolsep}{3pt} 
\begin{tabular}{p{35pt} p{75pt} p{75pt} p{30pt}} 
\hline
\textbf{Symbol} & \textbf{Env 1} & \textbf{Env 2} & \textbf{Unit} \\ 
\hline\hline
$q_s$ & $U(500, 3000)$ & $U(500, 3000)$ & mg/s \\ 
$\tau$ & $U(100, 1500)$ & $U(100, 1500)$ & s \\ 
$D$ & $U(1, 15)$ & $U(1, 15)$ & m$^2$/s \\ 
$V$ & $U(0, 5)$ & $U(0, 5)$ & m/s \\ 
$\chi$ & $U(0, 360)$ & $U(0, 360)$ & deg \\ 
$\sigma_{env}$ & 0.4 & 0.5 & mg/s \\ 
$\beta$ & 0.25 & 0.4 & - \\ 
\hline
\end{tabular}
\end{table}

\begin{table}[!t]
\setlength{\abovecaptionskip}{0.1pt}
\centering
\caption{Parameter lists and corresponding values}
\label{2}
\setlength{\tabcolsep}{6pt}
\renewcommand{\arraystretch}{1.05}
\begin{tabular}{p{174pt} p{40pt}}
\hline
\textbf{Parameter} & \textbf{Value} \\
\hline\hline
First fully connected layer size & 256 \\
Second fully connected layer size & 64 \\
Learning rate (actor, critic) & 0.0003 \\
Learning rate (curiosity network) & 0.00005 \\
Replay buffer size & 1,000,000 \\
Mini-batch size & 256 \\
Auxiliary reward weight ($\eta$) & 5 \\
Curiosity loss weighting factor $(\varepsilon)$ & 0.8 \\
Discount factor $(\gamma)$ & 0.99 \\
\hline
\end{tabular}
\end{table}
\begin{figure}[!t]
\centering
\includegraphics[width=0.48\textwidth]{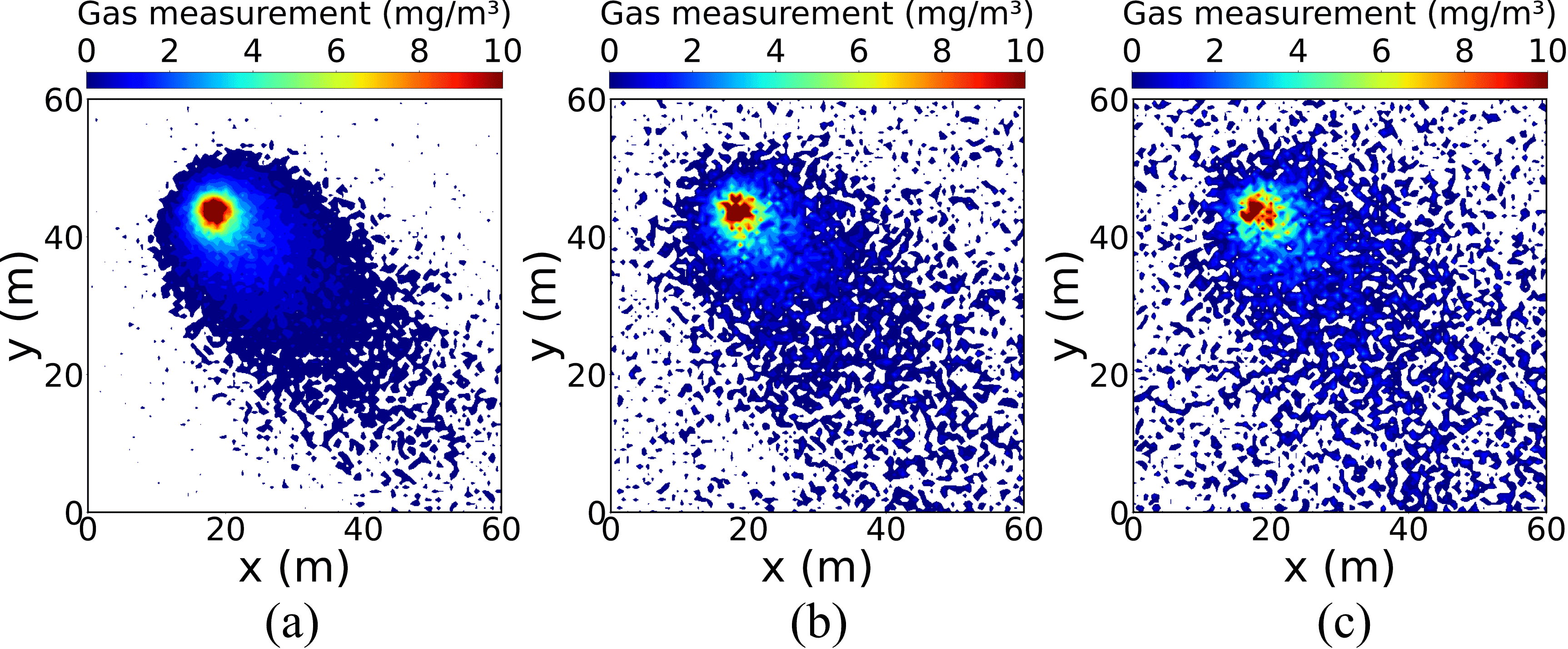} 
\vspace{-20pt}
\caption{Gas measurement maps under different environmental noise levels: (a) $\sigma_{{env}} = 0.2$, (b) $\sigma_{{env}} = 0.4$, and (c) $\sigma_{{env}} = 0.5$.}
\label{environment_noise}
\end{figure}
\section{Numerical Simulation}
\subsection{Simulation Setup}
The numerical simulations are conducted in two distinct environments, where gas properties and wind conditions are systematically varied, as summarized in Table~\ref{tab:parameters2}. Both environments 1 and 2 are configured under high-noise conditions, while environment~2 incorporates more challenging levels of environmental and sensor noise. Particularly, the environmental noise levels are set to 0.4 and 0.5, which are substantially higher than those used in prior studies (e.g., 0.2 in MVG-RDDPG~\cite{park2022mvgrddpg} and AGDC~\cite{aaai_ste}), resulting in more severe fluctuations in gas measurements. The effect of environmental noise is illustrated in Fig.~\ref{environment_noise}, which shows gas measurement maps under different noise levels. During training, the agent moves to the next sensing position with a fixed step size of \(2\,\text{m}\) within a \(60\,\text{m} \times 60\,\text{m}\) search area. A simulation terminates if the agent exceeds 300 steps or if the standard deviation of the particle filter with 2,000 particles falls below 0.1. At the end of each simulation, the source search is considered successful if the distance between the estimated and the true source is within \(3\,\text{m}\). All algorithms are trained for 80,000 episodes, and the training hyperparameters of the proposed method are listed in Table~\ref{2}. At the beginning of each episode, both the source and agent locations are randomly initialized, and the environmental parameters listed in Table~\ref{tab:parameters2} are randomly sampled from corresponding uniform distributions.

\subsection{Evaluation Metrics}
The trained policies are evaluated using two key metrics: success rate (SR) and mean travel distance (MTD). SR denotes the percentage of episodes in which the agent successfully estimates the source location. MTD is defined as the average travel distance computed over all episodes with successful source estimation. For each evaluation, 1,000 random scenarios that are not seen during training are executed. In addition, a stricter success criterion than in training is applied by requiring the distance between the true and estimated source locations to be within \(1\,\mathrm{m}\).

\begin{table}[t]
\setlength{\abovecaptionskip}{0.1pt}
\caption{Performance comparison with existing methods}
\label{performance comparison}
\centering
\begin{tabularx}{\columnwidth}{l *{4}{>{\centering\arraybackslash}X}}
\hline
\multirow{2}{*}[-0.5ex]{\centering\textbf{Method}} 
  & \multicolumn{2}{c}{\textbf{Env 1}} 
  & \multicolumn{2}{c}{\textbf{Env 2}} \\
\cline{2-3} \cline{4-5}
  & \rule{0pt}{2.3ex}\textbf{SR (\%)} & \textbf{MTD~(m)} 
  & \textbf{SR (\%)} & \textbf{MTD~(m)} \\
\hline\hline
Infotaxis [2]    & 87.6 & 155.6 & 83.0 & 169.0 \\
Entrotaxis [4]   & 85.4 & 142.3 & 82.7 & 149.2 \\
PC-DQN [12]      & 82.3 & 100.5 & 78.5 & 112.4 \\
MVG-RDDPG [13]   & 90.6 & 91.8  & 84.0 & 98.4  \\
AGDC [14]        & 86.4 & 98.4 & 82.8 & 104.3 \\
AID-RL [15]      & 73.9 & 120.4 & 63.4 & 138.2 \\
GMM IG-RL [16]  & 91.6 & 95.8  & 86.0 & 100.4  \\
CIG-RL (Ours)         & \textbf{98.3} & \textbf{80.2} & \textbf{95.0} & \textbf{86.2} \\
\hline
\end{tabularx}
\end{table}

\begin{figure}[!t]
\centering
\includegraphics[width=\columnwidth]{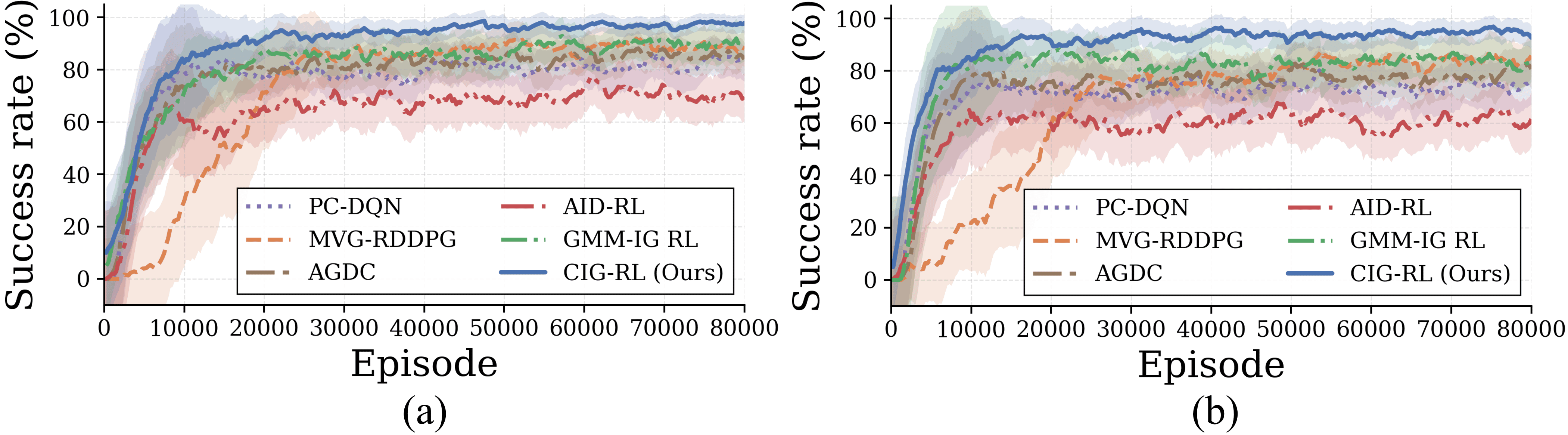} 
\vspace{-22pt}
\caption{Success rate curves during training in (a) environment 1 and (b) environment 2.}
\label{reward}
\end{figure}

\subsection{Performance Comparison with Existing Methods}
We compare our method with both information-theoretic and DRL-based approaches. The information-theoretic baselines include Infotaxis~\cite{vergassola2007infotaxis} and Entrotaxis~\cite{hutchinson2018entrotaxis}. The DRL-based baselines consist of PC-DQN~\cite{zhao2022pcdqn}, MVG-RDDPG~\cite{park2022mvgrddpg}, AGDC~\cite{aaai_ste}, AID-RL~\cite{li2023aidrl}, and GMM IG-RL~\cite{lee2025enhanced}. To ensure a fair comparison, all methods use the same gas dispersion and sensor models with 2,000 particles in the particle filter. In addition, all DRL-based methods are trained under identical noise settings listed in Table~\ref{tab:parameters2}.

\begin{figure}[!t]
\centering
\includegraphics[width=\columnwidth]{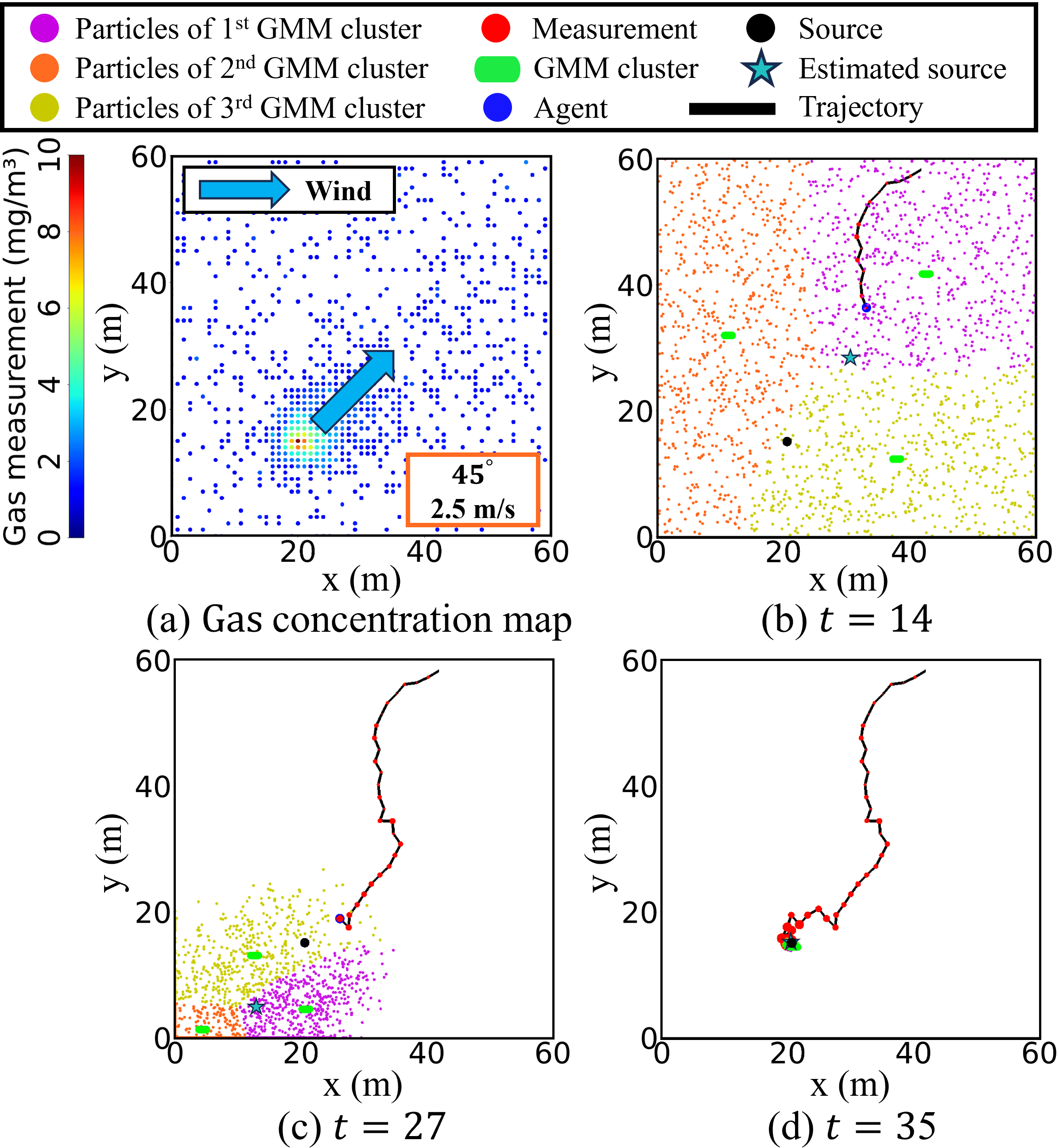}
\vspace{-18pt}
\caption{Illustrative run of CIG-RL in challenging noisy environment 2.}
\label{sample_trajectory}
\end{figure}

As shown in Table~\ref{performance comparison}, CIG-RL demonstrates the most robust performance in both environments, achieving SR of 98.3\% and 95.0\%, respectively. The training curves in Fig.~\ref{reward} show that CIG-RL achieves the lowest SR variability and the highest SR among DRL baselines, indicating that CIG-RL learns a robust search policy under noisy conditions. Sample trajectories of CIG-RL in environment~2 are presented in Fig.~\ref{sample_trajectory}. Larger red circles denote higher gas concentrations.

\subsection{Ablation Study}
\subsubsection{Ablation Analysis of CIG-RL}
An ablation analysis is conducted in both environments to evaluate the contributions of the proposed components in CIG-RL. As shown in Table~\ref{ablation}, removing both the proposed exploration strategy and the active perception reward significantly degrades performance. This result suggests that insufficient belief space exploration and less informative reward guidance can hinder robust policy learning under highly noisy measurements. When curiosity and the active perception reward are removed individually, the performance improves compared with removing both components, but SR remains lower than that of CIG-RL. These results demonstrate that coupling belief space exploration with active perception reward guidance is beneficial for robust STE in highly noisy environments.

In addition, the results in Table~\ref{ablation} show that MTD increases when the particle filter variance is not incorporated in the active perception reward. This highlights that the uncertainty-adaptive adjustment between the mutual information term and the distance term plays a crucial role in improving search efficiency. Furthermore, CIG-RL, which uses the decoupled sampling strategy, outperforms the variant that samples the auxiliary reward from the replay buffer. Since reusing samples with previously high curiosity rewards in later training stages can encourage excessive exploration, this can lead to inefficient source search behavior.
\subsubsection{Sensitivity Analysis of CIG-RL}
Next, we analyze the sensitivity to the weighting factor \( \eta \) between the extrinsic and auxiliary rewards. As outlined in Table~\ref{sensivity}, when \( \eta \) is too small, SR degrades since the curiosity becomes insufficient to promote exploration of novel belief transitions. In contrast, when \( \eta \) becomes larger than 5, the auxiliary reward leads to excessive exploration in the early stage of training, consequently increasing the MTD.
\subsubsection{Analysis of Curiosity in CIG-RL}
\begin{figure}[!t]
\centering
\includegraphics[width=\columnwidth]{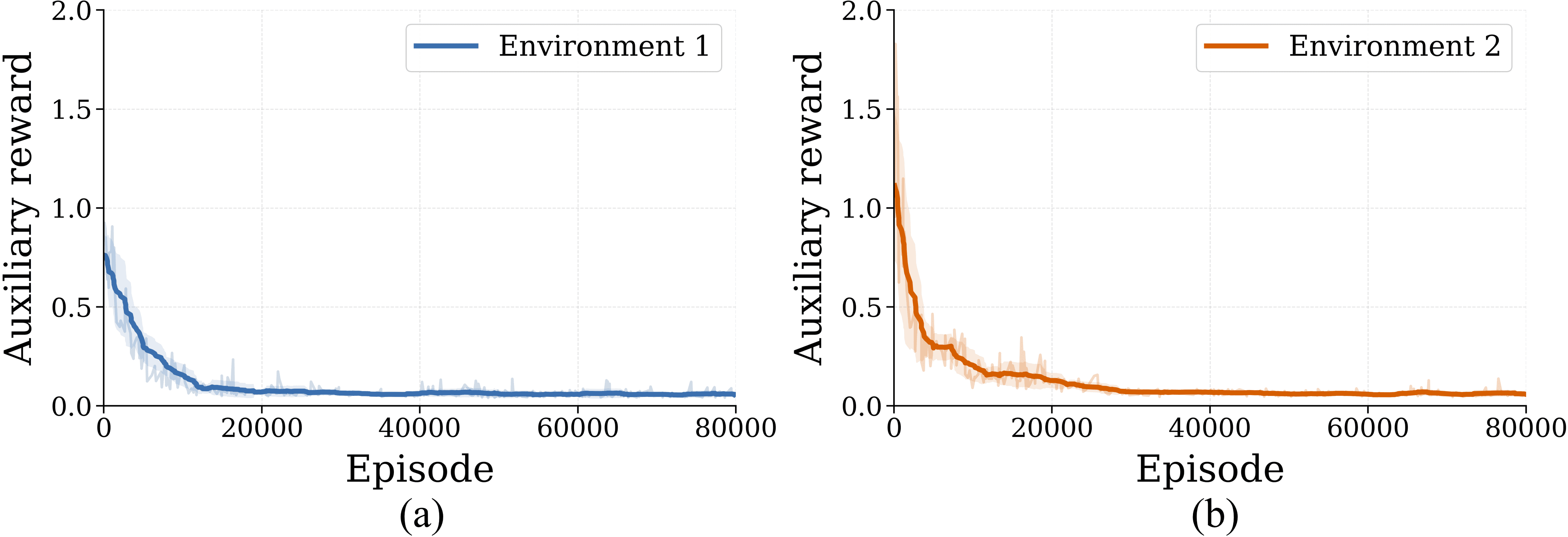} 
\vspace{-22pt}
\caption{Auxiliary reward curves of CIG-RL during training under two high-noise settings: 
(a) environment 1 ($\sigma_{{env}}=0.4$, $\beta=0.25$) and 
(b) environment 2 ($\sigma_{{env}}=0.5$, $\beta=0.4$).}
\label{forwardloss}
\end{figure}

Fig.~\ref{forwardloss} shows the curiosity auxiliary reward, defined as the forward network loss, during training under two high-noise settings. The auxiliary reward is initially high but consistently decreases as the agent explores belief transitions and the forward network is trained, converging to a low value in both environments. This demonstrates stable convergence of the proposed belief-based curiosity despite noisy measurements.

\begin{table}[!t]
\setlength{\abovecaptionskip}{0.1pt}
\caption{Ablation study of CIG-RL in two environments}
\label{ablation}
\centering
\begin{tabularx}{\columnwidth}{l *{4}{>{\centering\arraybackslash}X}}
\hline
\multirow{2}{*}[-0.5ex]{\centering\textbf{Method}}
  & \multicolumn{2}{c}{\textbf{Env 1}}
  & \multicolumn{2}{c}{\textbf{Env 2}} \\
\cline{2-3} \cline{4-5}
  & \rule{0pt}{2.3ex}\textbf{SR (\%)} & \textbf{MTD (m)}
  & \textbf{SR (\%)} & \textbf{MTD (m)} \\
\hline\hline
w/o curiosity \& active reward       & 86.0 & 96.1 & 82.5 & 103.2 \\
w/o curiosity     & 91.8 & 88.5 & 88.3 & 97.5 \\
w/o active reward     & 93.0 & 84.3 & 89.7 & 92.6 \\ 
w/o particle filter variance     & 96.1 & 88.8 & 94.3 & 93.4 \\
w/o decoupled sampling     & 95.4 & 90.4 & 92.8 & 96.2 \\
CIG-RL (Ours)   & \textbf{98.3} & \textbf{80.2} & \textbf{95.0} & \textbf{86.2} \\
\hline
\end{tabularx}
\end{table}

\begin{table}[!t]
\setlength{\abovecaptionskip}{0.1pt}
\caption{Performance evaluation of auxiliary reward sensitivity}
\label{sensivity}
\centering
\begin{tabularx}{\columnwidth}{l *{4}{>{\centering\arraybackslash}X}}
\hline
\multirow{2}{*}[-0.5ex]{\centering\textbf{Method}}
  & \multicolumn{2}{c}{\textbf{Env 1}}
  & \multicolumn{2}{c}{\textbf{Env 2}} \\
\cline{2-3} \cline{4-5}
  & \rule{0pt}{2.3ex}\textbf{SR (\%)} & \textbf{MTD (m)}
  & \textbf{SR (\%)} & \textbf{MTD (m)} \\
\hline\hline
\( \eta = 0.5 \)      & 95.0 & 83.0 & 92.1 & 88.8 \\
\( \eta = 1 \)    & 95.6 & 81.6 & 93.0 & 86.4 \\
\( \eta = 2.5 \)    & 97.1 & \textbf{78.4} & 93.3 & \textbf{85.6} \\
\( \eta = 5 \)    & \textbf{98.3} & 80.2 & \textbf{95.0} & 86.2 \\
\( \eta = 10 \)  & 97.6 & 85.8 & 94.6 & 91.7 \\
\hline
\end{tabularx}
\end{table}

\section{Real-World Experiments}
\begin{figure}[!t]
\centering
\includegraphics[width=0.47\textwidth]{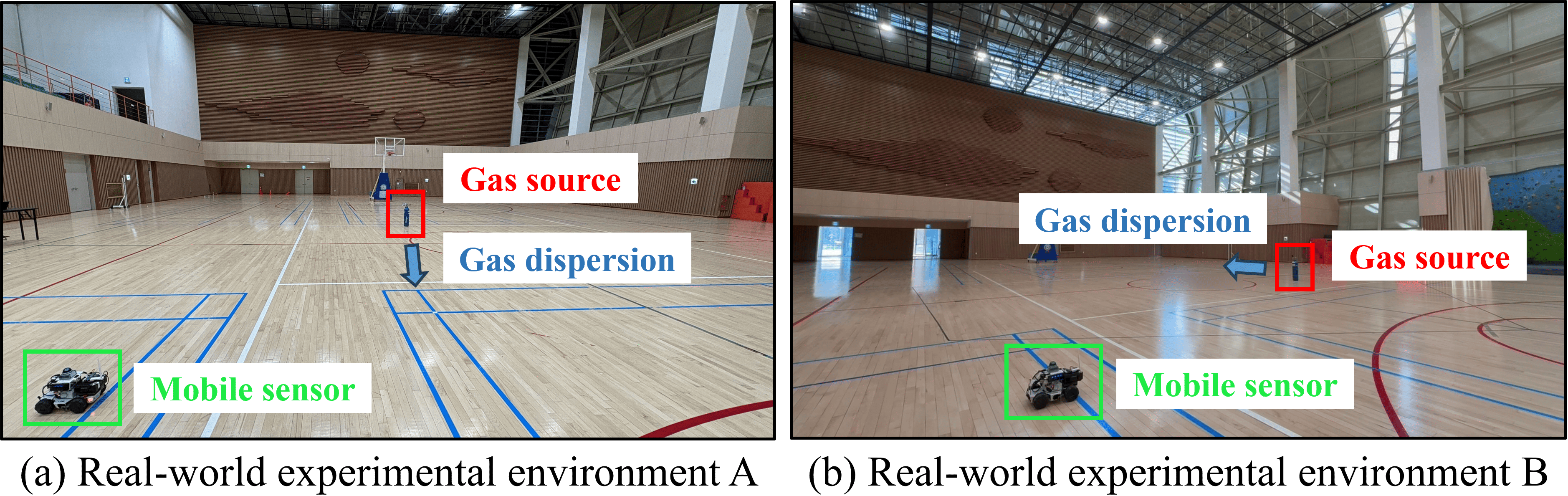} 
\vspace{-10pt}
\caption{Overview of the real-world experimental environments.}
\label{setup}
\end{figure}

As illustrated in Fig.~\ref{experiment1}, a \(CO_2\) tank is used to generate gas dispersion for the real-world experiments. An AgileX Limo equipped with a SEN15112 gas sensor serves as the mobile sensing platform, with onboard computation performed by an NVIDIA Jetson Xavier NX. A Livox Mid-360 LiDAR provides environmental perception, while FAST-LIO2~\cite{xu2022fast} is used for mobile sensor localization. Each algorithm is tested 10 times in an \(8\,\text{m} \times 8\,\text{m}\) gym, where the mobile sensor starts at \(x=0\,\text{m}\) and \(y=0\,\text{m}\) and moves \(0.5\,\text{m}\) per step. Each trial terminates when the particle filter variance falls below 0.1, and success of source search is declared if the distance between the estimated and true source is within \(0.5\,\text{m}\). In environment~A, shown in Fig.~\ref{setup}(a), the source is located at \(x=4\,\text{m}\) and \(y=8\,\text{m}\) and released toward the negative \(y\)-direction. In environment~B, shown in Fig.~\ref{setup}(b), it is at \(x=8\,\text{m}\) and \(y=4\,\text{m}\) and released along a \(45^\circ\) upward-left direction, resulting in a more challenging scenario where the plume is advected far from the mobile sensor, leading to sparse measurements. To ensure steady gas dispersion, \(CO_2\) is released for 4 minutes before each trial. Measurements are collected 5 seconds after the mobile sensor reaches each sensing position considering the sensor response time, and the gym windows remain open to introduce sufficient external airflow.

\begin{table}[t!]
\setlength{\abovecaptionskip}{0.1pt}
\caption{Results of real-world experiments}
\label{realworld}
\centering
\begin{tabularx}{\columnwidth}{l *{6}{>{\centering\arraybackslash}X}}
\hline
\multirow{2}{*}[-0.5ex]{\centering\textbf{Method}}
  & \multicolumn{3}{c}{\textbf{Env A}}
  & \multicolumn{3}{c}{\textbf{Env B}} \\
\cline{2-4} \cline{5-7}
  & \rule{0pt}{2.3ex}\textbf{SR (\%)} 
  & \textbf{SN} 
  & \textbf{CT (ms)}
  & \textbf{SR (\%)} 
  & \textbf{SN} 
  & \textbf{CT (ms)} \\
\hline\hline
Infotaxis       & 70  & 49.3 & 420.3 & 60 & 61.3 & 425.4 \\
MVG-RDDPG       & 80  & 34.8 & \textbf{80.4} & 60 & 48.6 & \textbf{79.8} \\
CIG-RL (Ours)          & \textbf{100} & \textbf{26.4} & 82.6 & \textbf{90} & \textbf{30.8} & 83.8 \\
\hline
\end{tabularx}
\end{table}

In the real-world experiments, we compare Infotaxis, MVG-RDDPG, and CIG-RL, while additionally evaluating the average computation time (CT) for one-step decision making. As shown in Table~\ref{realworld}, Infotaxis exhibits the highest CT in both environments because it calculates the expected entropy reduction for four action candidates by considering possible future measurements. In contrast, MVG-RDDPG and CIG-RL show substantially lower CT by learning policies through DRL. Although CIG-RL shows slightly higher CT than MVG-RDDPG, it shows more robust performance in terms of SR and step number (SN). Fig.~\ref{experiment1} shows that CIG-RL achieves a more efficient search path toward the source than Infotaxis and MVG-RDDPG. Particularly, CIG-RL approaches the source without the oscillatory movements observed in Infotaxis and MVG-RDDPG, resulting in fewer steps to particle filter convergence. A video of the real-world experiments is available at [URL will be added upon acceptance].

\section{Conclusions and future work}\label{conclusion}
In this study, we propose a curiosity-driven information-guided reinforcement learning framework for robust source search in noisy environments. The proposed framework couples curiosity-driven belief space exploration with an uncertainty-adaptive active perception mechanism, enabling the agent to efficiently obtain informative measurements while maintaining robust policy updates under noisy conditions. Simulations under highly noisy environmental conditions and real-world experiments show that CIG-RL outperforms existing approaches in terms of success rate and search efficiency, while retaining practical computational cost for real-time deployment. For future work, we plan to extend the framework to multi-agent systems to further improve search efficiency and scalability in more complex environments.

\addtolength{\textheight}{-12cm}   

\bibliographystyle{IEEEtran}
\bibliography{references.bib}

\end{document}